\documentclass[letterpaper, 10 pt, conference]{ieeeconf}

\IEEEoverridecommandlockouts

\usepackage{amsmath,amssymb,amsfonts}
\usepackage{graphicx}
\usepackage{booktabs}
\usepackage{subcaption}
\usepackage[table]{xcolor}
\usepackage{hyperref}
\usepackage{makecell}
\usepackage{cite}

\begin{document}

\title{Seeing Where to Deploy: Metric RGB-Based Traversability Analysis\\for Aerial-to-Ground Hidden Space Inspection}

\author{
Seoyoung Lee$^{1}$, Shaekh Mohammad Shithil$^{2}$, Durgakant Pushp$^{2}$, Lantao Liu$^{2}$, Zhangyang Wang$^{1}$\thanks{$^{1}$The University of Texas at Austin, USA. \{seoyounglee, atlaswang\}@utexas.edu.
$^{2}$Indiana University, Bloomington, USA. \{sshithil, dpushp, lantao\}@iu.edu.
}
}

\maketitle

\begin{abstract}

Inspection of confined infrastructure such as culverts often requires accessing hidden spaces whose entrances are reachable primarily from elevated viewpoints. Aerial–ground cooperation enables a UAV to deploy a compact UGV for interior exploration, but selecting a suitable deployment region from aerial observations requires metric terrain reasoning involving scale ambiguity, reconstruction uncertainty, and terrain semantics. We present a metric RGB-based geometric–semantic reconstruction and traversability analysis framework for aerial-to-ground hidden space inspection. A feed-forward multi-view RGB reconstruction backbone produces dense geometry, while temporally consistent semantic segmentation yields a 3D semantic map. To enable deployment-relevant measurements without requiring LiDAR-based dense mapping, we introduce an embodied motion prior that recovers metric scale by aligning predicted camera motion with onboard platform egomotion. From the metrically grounded reconstruction, we construct a confidence-aware geometric–semantic traversability map and evaluate candidate deployment zones under reachability constraints. Experiments on a tethered UAV–UGV platform demonstrate reliable deployment-zone identification in hidden space scenarios. [Project page: \url{https://seoyoung1215.github.io/seeing-where-to-deploy/}]

\end{abstract}

\section{Introduction}
\label{sec:intro}

Robotic inspection of infrastructure and confined environments remains a critical challenge in field robotics. Many inspection targets such as culverts, drainage systems, ventilation shafts, and disaster voids are partially occluded, geometrically constrained, and difficult to access from ground level~\cite{murphy2004trial, lee2025cipher}. We refer to such environments as \emph{hidden spaces}: regions whose entrances are visible only from elevated viewpoints and whose interiors are initially unobservable and uncertain~\cite{pushp2026miniugv2}. Reliable access to these spaces is essential for infrastructure maintenance, environmental monitoring, and search-and-rescue operations.
In hidden-space scenarios, access is fundamentally aerial-first. An unmanned ground vehicle (UGV) alone cannot reach elevated entrances due to obstacles or terrain discontinuities. While aerial platforms (UAVs) can localize and easily access potential entry points of hidden space from overhead viewpoints, they are limited in endurance and interaction capability. This asymmetry motivates aerial–ground cooperation in which the UAV serves as the primary access platform and deploys a compact UGV for interior exploration~\cite{pushp2022uav}.

Aerial deployment of a ground robot is
a safety-critical decision problem under perception uncertainty and limited onboard computation. Once a UGV is lowered and released, recovery may be difficult or infeasible. Thus, deployment must ensure a physically suitable landing region and a feasible path to the hidden entrance despite perception uncertainty. This requires metric validation of the deployment zone and its surrounding terrain prior to release.

Recent advances in feed-forward RGB-based multi-view reconstruction enable dense geometric prediction without iterative bundle adjustment or heavy sensing payloads~\cite{wang2025vggt}. However, such methods recover scene structure only up to an unknown global similarity transformation, leaving absolute scale ambiguous. For deployment-critical tasks, metric quantities such as obstacle clearance, slope limits, and terrain height variation directly determine physical safety margins.
In hidden-space inspection scenarios, the UAV typically operates from predominantly top-down viewpoints, where limited triangulation baselines degrade reconstruction accuracy for ground-level discontinuities and vertical structures, while wind-induced motion during hover introduces registration noise that can produce duplicated or smeared surface estimates. These geometric uncertainties arise precisely in the regions where accurate terrain characterization is most critical. Addressing them under real-time and memory constraints is essential for safe aerial-to-ground deployment.

Moreover, geometric reasoning alone is insufficient for reliable traversability assessment. Terrain materials such as compact soil, grass, loose gravel, or mud may exhibit similar geometric profiles while having very different support and traction properties. Therefore, deployment-zone selection must incorporate semantic information and verify that a feasible path exists from the landing region to the hidden entrance while avoiding hazardous terrain.
Existing aerial mapping and reconstruction approaches primarily emphasize geometric fidelity or dense scene modeling~\cite{wang2025vggt, yang2025fast3r}, but do not explicitly address deployment-oriented metric grounding, reconstruction confidence attenuation, and reachability-aware zone evaluation under perception uncertainty.

To overcome these limitations, we propose an RGB-based geometric–semantic reconstruction pipeline with embodied metric grounding for aerial-to-ground hidden-space inspection. Our framework leverages feed-forward multi-view reconstruction for dense geometry, integrates temporally consistent semantic segmentation, and recovers metric scale using onboard egomotion as a physical prior. From the metrically grounded semantic reconstruction, we derive a confidence-aware bird's-eye-view traversability representation and evaluate candidate deployment zones under explicit reachability constraints, prior to release.

The main contributions of this work are:
(1) a feed-forward pipeline that coherently merges geometric and semantic 3D reconstruction;
(2) an embodied metric-grounding strategy using onboard egomotion to enable deployment-relevant reasoning from RGB-based reconstruction;
(3) an aerial traversability estimation framework that fuses geometric terrain features and semantic information to identify safe deployment zones for ground robots.
The proposed method is validated on field-collected data from a UAV–miniUGV inspection platform in hidden space scenarios.

\section{Related Work}
\label{sec:related}

Although progress in sensing technologies, robotic platforms, and AI-driven analysis has enabled early developments in automated inspection technologies~\cite{
gibb2018nondestructive, sanchez2019robotic, la2019automated, li2019cable}, many existing approaches still rely heavily on manual analysis of collected data, resulting in substantial time and human effort. Thus, developing a fully automated inspection and management framework remains an important open challenge.

Accurate 3D reconstruction is a prerequisite for safe hidden space exploration, as geometry directly informs structural assessment and traversability estimation. However, existing reconstruction pipelines face significant limitations under deployment constraints. Structure-from-Motion (SfM)-based methods~\cite{schoenberger2016sfm,
mildenhall2020nerf, kerbl3Dgaussians} often degrade and fail with insufficient viewpoint diversity or parallax, leading to unstable pose estimation and error propagation into dense reconstruction. Learning-based multi-view stereo approaches have also emerged~\cite{leroy2024grounding, fan2024large, lee2025cipher, yang2025fast3r, chen2025easi3r}, some enabling high-fidelity reconstructions, but often fail at purely top-down views or are not efficient or lightweight to be incorporated in UAV inspection systems. While LiDAR sensing provides metric depth, it introduces practical challenges for aerial deployment, including payload weight, energy consumption, and reduced accuracy at higher elevations or grazing angles. These constraints motivate a lightweight, vision-driven alternative suitable for UAV-based systems.

Terrain traversability estimation is fundamental for autonomous navigation in unstructured environments. Early approaches relied on geometric terrain representations derived from range sensing, such as elevation maps that capture surface height and uncertainty to estimate slope, roughness, and obstacle presence~\cite{Fankhauser2018}. More recent work incorporates visual perception to infer terrain properties directly from images using learning-based methods, enabling robots to predict traversability from visual cues through self-supervised learning~\cite{Wellhausen2019}. Recent advances further leverage large pre-trained visual models and online self-supervision to rapidly adapt traversability prediction across diverse outdoor environments~\cite{Mattamala2025}. In parallel, semantic perception methods improve scene understanding by classifying terrain categories and enabling more robust navigation in challenging environments such as construction sites or off-road terrain~\cite{khan2025afrda,Guan2022TNS}.
However, many of these approaches assume ground-level sensing or metric depth measurements, whereas aerial deployment requires terrain reasoning from predominantly top-down RGB observations where dense geometric reconstruction must be inferred without dedicated depth sensors.

\section{Research Background and Preliminaries}

This section outlines the core components that provide the foundational building blocks for our proposed framework.

\begin{figure*}[t]
  \centering
  \includegraphics[width=0.98\linewidth]{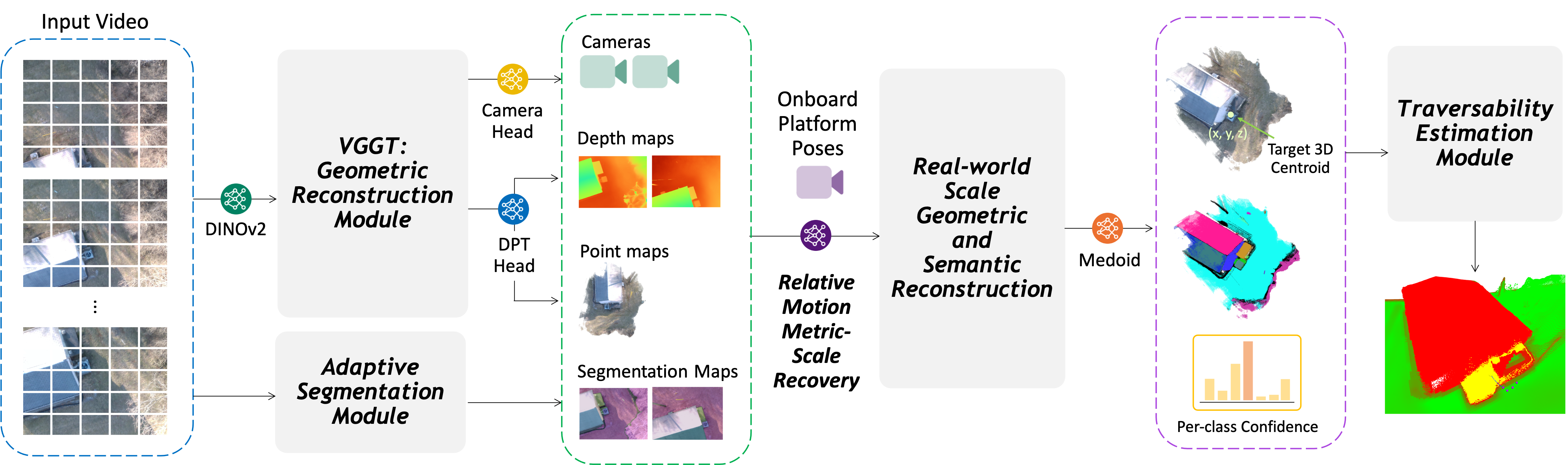}
  \caption{\small \textbf{Overview of proposed pipeline.} From a UAV-captured top-down RGB video, VGGT predicts camera parameters and depth, which yield dense 3D geometry via depth unprojection. In parallel, adaptive segmentation generates temporally consistent instance masks that are lifted into 3D to produce a geometry-aligned semantic reconstruction. Onboard egomotion recovers metric scale for the reconstruction through motion-consistent alignment, enabling deployment-relevant target extraction for traversability analysis and deployment planning. \vspace{-10pt}}
  \label{fig:method_perception}
\end{figure*}

\subsection{Visual Geometry Grounded Transformer (VGGT)
}

VGGT~\cite{wang2025vggt} directly predicts camera  and dense geometric parameters from a set of uncalibrated RGB images in a single forward pass, eliminating iterative optimization (e.g., bundle adjustment in SfM pipelines) and explicit post-processing.

Given $N$ RGB images observing a static scene, $\mathcal{I} = \{ I_i \}_{i=1}^{N}$ with $I_i \in \mathbb{R}^{3 \times H \times W}$, each image is patchified using a DINOv2-based backbone to produce visual tokens $\mathbf{t}_i \in \mathbb{R}^{K \times C}$. A learnable camera token is appended to each frame, with a distinct token assigned to the first frame to define the reference coordinate frame. The concatenated tokens are processed by $L$ alternating self-attention layers that switch between frame-wise attention (within each image) and global attention (across images), enabling multi-view geometric reasoning without explicit geometric constraints.

From the refined tokens, a camera head predicts per-frame intrinsics and extrinsics $\mathbf{g}_i = (\mathbf{R}_i, \mathbf{t}_i, \mathbf{f}_i)$, where $\mathbf{R}_i \in SO(3)$, $\mathbf{t}_i \in \mathbb{R}^3$, and $\mathbf{f}_i$ are focal parameters. The first camera defines the world frame with $\mathbf{R}_1 = \mathbf{I}$ and $\mathbf{t}_1 = \mathbf{0}$. A Dense Prediction Transformer (DPT) head predicts per-frame depth maps $\mathbf{D}_i \in \mathbb{R}^{H \times W}$ and corresponding confidence maps $\mathbf{Q}_i$.

VGGT is trained end-to-end with a multi-task loss combining camera, depth, point-map, and tracking supervision. This feed-forward formulation enables efficient reconstruction from RGB sequences without known calibration, experiment control, or bundle adjustment, and generalizes well to diverse viewpoints, including top-down exploration scenarios.

\subsection{3D Semantic Segmentation Mapping}

To project semantic information into 3D, we require instance masks that are temporally consistent across a video. Frame-wise segmentation alone produces fragmented or identity-inconsistent masks, leading to unstable 3D label projection. This issue is particularly pronounced in inspection scenarios, where a moving viewpoint reveals previously unseen structural regions.

Given an RGB frame $I_t$, SAM predicts a set of candidate masks
$
\mathcal{M}^{\text{SAM}}_t=\{M_t^k\}_{k=1}^{K_t},
$
where $M_t^k \in \{0,1\}^{H\times W}$.
While SAM provides high-quality per-frame instance proposals, it does not enforce temporal identity consistency. Recent extensions such as SAM~2 enable mask propagation and tracking over time using motion cues. However, these approaches primarily maintain consistency for previously discovered objects and do not inherently address the discovery of newly visible regions that appear as the viewpoint changes.
In our framework, temporally consistent instance masks serve as the semantic input that is later projected to the reconstructed geometry for deployment-oriented terrain reasoning.

\section{Proposed Framework}

The proposed framework shown in Figure~\ref{fig:method_perception} integrates perception, metric grounding, and deployment reasoning into a unified pipeline for aerial-to-ground deployment. Given an RGB video stream collected from a hovering UAV with exclusively top-down views, we first generate a dense multi-view geometric reconstruction and temporally consistent semantic map (Sec.~\ref{sec:sem}). We then enforce metric consistency using an embodied motion prior derived from onboard egomotion, yielding navigation-grade 3D structure suitable for deployment-critical reasoning (Sec.~\ref{sec:metric_scale}). From the metrically grounded reconstruction, we construct a bird's-eye-view (BEV) traversability representation that fuses geometric cues, semantic compatibility, and reconstruction confidence attenuation. Finally, candidate deployment zones are evaluated under path feasibility criteria prior to physical release (Sec.~\ref{sec:trav-map}).

\subsection{Semantic and Geometric Reconstruction}

\subsubsection{RGB-based 3D Geometric-Semantic Reconstruction}
\label{sec:sem}

To enable efficient multi-view 3D reconstruction from RGB images, we adopt VGGT as our
reconstruction backbone.

In our formulation, we omit the 3D point DPT head and instead obtain 3D points by inverse-projecting the predicted depth using the estimated camera parameters, which yields more accurate geometry. Specifically,
\begin{equation}
\mathbf{X}_i(u,v) = \mathbf{R}_i^{-1} \left( \mathbf{K}_i^{-1} [u, v, 1]^\top \mathbf{D}_i(u,v) - \mathbf{t}_i \right),
\end{equation}
where $(\mathbf{R}_i,\mathbf{t}_i)$ denotes a world-to-camera transform $\mathbf{x}_i^c=\mathbf{R}_i\mathbf{X}+\mathbf{t}_i$, and $\mathbf{K}_i$ is constructed from $\mathbf{f}_i$. The reconstructed point cloud $\mathbf{X}_S$ is expressed in the first-camera coordinates.

For segmentation, we apply overlap-based suppression (Mask-NMS) to remove redundant proposals and retain high-quality masks. At the initial keyframe, each retained mask is assigned a unique instance ID and used as a dense prompt for SAM 2 to initiate video-level propagation.

Instead of independently segmenting every frame, we adopt an incremental keyframe-driven discovery strategy for enhanced efficiency. Let $\Omega_t$ denote the pixels covered by propagated masks in frame $t$, and define uncovered area ratio
\begin{equation}
\rho_t = 1 - \frac{|\Omega_t|}{H W}.
\end{equation}
When $\rho_t$ exceeds a threshold $\delta$ relative to previous keyframes, indicating newly observed geometry entering the field of view, we trigger a new keyframe. SAM is re-run on that frame, and masks covering unexplained regions are added as new instances. All instance masks are then propagated across frames via SAM 2 to update the temporally consistent set of masks
$\hat{\mathcal{M}}^{k}=\{\hat{M}_t^{k}\}_{t=1}^{T}$.

This adaptive expansion ties instance discovery to scene coverage changes induced by camera motion. The tracked instance set grows only when the uncovered-area ratio indicates newly visible regions, avoiding redundant per-frame segmentation while preserving global temporal consistency under resource constraints of real-time robotic perception.

Finally, each pixel $(u,v)$ is back-projected to $\mathbf{X}_t(u,v)\in\mathbb{R}^3$ with VGGT-predicted depth and camera parameters. Let $\mathcal{K}_t(u,v)=\{k:\hat{M}_t^k(u,v)=1\}$ denote mask indices covering $(u,v)$. We assign $\ell(\mathbf{X}_t(u,v))$ the mask label if $|\mathcal{K}_t(u,v)|=1$, a composite overlap label if $|\mathcal{K}_t(u,v)|>1$, and leave it unlabeled otherwise. Aggregating labeled points across frames yields a geometry-aligned, instance-consistent 3D semantic reconstruction from RGB input.

\subsubsection{Embodied Metric-Scale Geometric Reconstruction}
\label{sec:metric_scale}

Feed-forward RGB reconstruction recovers scene geometry only up to an unknown global similarity transformation due to scale ambiguity in monocular perspective projection.

Rather than relying on dense LiDAR mapping, external map alignment, or retraining with metric supervision, we metrically ground the RGB-based reconstruction using an embodied motion prior from onboard platform egomotion. The formulation is agnostic to the egomotion source and can use various onboard state estimation approaches, including VIO, GPS fusion, and FAST-LIO. The resulting metric inter-frame displacements provide motion-consistency constraints for recovering global scale.

Let $\mathbf{C}_i^{v} \in \mathbb{R}^3$ denote the camera centers predicted by VGGT in its internal reconstruction frame, and let $\mathbf{C}_i^{p} \in \mathbb{R}^3$ denote corresponding platform camera centers obtained from onboard state estimation. Since the global reference frame of the platform trajectory may be arbitrary, we estimate metric scale using relative motion consistency.

Inter-frame displacements are defined as
\begin{equation}
\Delta \mathbf{C}_{i,\eta}^{v}
=
\mathbf{C}_{i+\eta}^{v}-\mathbf{C}_{i}^{v},
\qquad
\Delta \mathbf{C}_{i,\eta}^{p}
=
\mathbf{C}_{i+\eta}^{p}-\mathbf{C}_{i}^{p}
\end{equation}
where multiple temporal strides $\eta\in\mathcal{A}=\{1,2,4,8\}$ are used to improve robustness against noise and drift.

We estimate a similarity transformation in $\mathrm{Sim}(3)$ parameterized by scale $s \in \mathbb{R}^{+}$, rotation $\mathbf{R} \in \mathrm{SO}(3)$, and translation $\mathbf{t} \in \mathbb{R}^3$ such that platform-consistent motion is satisfied:
\begin{equation}
\Delta \mathbf{C}_{i,\eta}^{p}
\approx
s\,\mathbf{R}\,\Delta \mathbf{C}_{i,\eta}^{v}.
\end{equation}
The optimal parameters are obtained via closed-form Umeyama alignment of relative displacements,
\begin{equation}
(s,\mathbf{R})
=
\arg\min_{s,\mathbf{R}}
\sum_{\eta\in\mathcal{A}}
\sum_{i=1}^{N-\eta}
\left\|
\Delta \mathbf{C}_{i,\eta}^{p}
-
s\,\mathbf{R}\,\Delta \mathbf{C}_{i,\eta}^{v}
\right\|_2^2.
\end{equation}
Translation is recovered by anchoring the first camera center,
\begin{equation}
\mathbf{t}
=
\mathbf{C}_1^{p}
-
s\,\mathbf{R}\,\mathbf{C}_1^{v}.
\end{equation}

The estimated similarity transform is applied consistently to all reconstructed geometry and camera poses. For a camera-to-world pose $(\mathbf{R}_{wc}, \mathbf{t}_{wc})$, the metrically grounded pose becomes
\begin{equation}
\mathbf{R}_{wc}' = \mathbf{R}\,\mathbf{R}_{wc},
\qquad
\mathbf{t}_{wc}' = s\,\mathbf{R}\,\mathbf{t}_{wc} + \mathbf{t}.
\end{equation}
When geometry is recovered via depth unprojection, predicted depths are additionally scaled by $s$ to maintain consistency between camera motion and reconstructed structure.

Unlike post hoc map alignment, we do not align to an external reference map, perform loop closure or conduct global SLAM refinement. Instead, we enforce motion consistency between learned geometric predictions and physically realized platform displacement. The reconstruction is therefore metrically grounded solely through lightweight egomotion signals that are routinely available on UAV platforms.

From a deployment perspective, this embodied grounding is critical. Deployment feasibility depends on metric quantities such as obstacle clearance distance, terrain slope thresholds, and residual energy margins. By recovering global scale without
dense LiDAR terrain mapping or iterative bundle adjustment, the proposed method enables navigation-grade semantic reconstruction within the payload and computational constraints of aerial platforms.

Metric grounding also enables extraction of deployment-relevant semantic primitives directly in 3D space.
For a semantic class $c$, we aggregate all reconstructed points belonging to the corresponding instance and compute a representative 3D coordinate using a medoid estimate, which is guaranteed to lie within the segmented region. When medoid computation becomes computationally prohibitive, the centroid of the associated point set is used as a fallback approximation. In addition, a class confidence value is computed by averaging per-pixel semantic probabilities projected into 3D space, yielding a continuous confidence measure reflecting segmentation reliability. This formulation supports overlapping semantic hypotheses by preserving confidence scores for multiple classes when present. The resulting
metrically grounded target centroids and semantic confidence values enable downstream deployment reasoning, target localization, and risk-aware decision making directly from RGB-based perception.

\subsection{Traversability Mapping from Aerial View}
\label{sec:trav-map}
Autonomous UAV-based deployment requires assessing terrain suitability prior to physical interaction. In our framework, the reconstruction pipeline
generates a semantically labeled 3D scene from aerial RGB observations and the UAV evaluates candidate deployment regions based on traversability. Let the reconstructed scene be represented as a set of 3D points $\mathcal{C} = \{(x_i, y_i, z_i, c_i, \pi_i)\}_{i=1}^M,$ where $(x_i, y_i, z_i)$ are world-frame coordinates, $c_i \in \{1,\dots,N\}$ denotes the semantic class label, and $\pi_i \in [0,1]$ represents reconstruction confidence. To enable spatial reasoning, the scene is projected onto a bird's-eye-view (BEV) grid with resolution $d_{\mathrm{res}}$. For each grid cell $g$, we define $\mathcal{P}(g)$ as the set of reconstructed points whose projections fall within that cell. This BEV discretization provides the spatial basis for geometric and semantic traversability estimation.

\subsubsection{Semantic Information}

For each BEV grid cell $g$, we derive semantic attributes from the reconstructed point set. Let $\mathcal{P}(g)$ denote the set of reconstructed points whose projections fall within cell $g$. The dominant semantic label is obtained via majority voting:
\begin{equation}
\hat{c}_{\mathrm{sem}}(g)
=
\arg\max_{c \in \{1, \dots, N\}}
\sum_{p_i \in \mathcal{P}(g)}
\mathbf{1}[c_i = c],
\end{equation}
where $c_i$ is the semantic class of point $p_i$ and $\mathbf{1}[\cdot]$ is the indicator function. To account for reconstruction reliability, we compute the mean confidence per cell:
\begin{equation}
p_{\text{conf}}(g)
=
\frac{1}{|\mathcal{P}(g)|}
\sum_{p_i \in \mathcal{P}(g)}
\pi_i,
\end{equation}
where $\pi_i \in [0,1]$ denotes the confidence associated with each 3D point. The quantities $\hat{c}_{\mathrm{sem}}(g)$ and $p_{\text{conf}}(g)$ provide terrain categorization and reliability information used in the subsequent traversability formulation.

\subsubsection{Geometric Information}
\label{sec:geo_info}

Height, slope, roughness, and clearance features are extracted on the BEV grid to characterize terrain structure
for UAV landing and UGV deployment.

\noindent
\textbf{Height Estimation.}
Each grid cell $g$ is represented as a 3D point
$\mathbf{p}_g = (x_g, y_g, \bar{h}(g))$,
with cell center $(x_g,y_g)$ and mean terrain height $\bar{h}(g)$ computed from points in $\mathcal{P}(g)$:
\begin{equation}
\bar{h}(g)
=
\frac{1}{|\mathcal{P}(g)|}
\sum_{p_i \in \mathcal{P}(g)} z_i.
\label{eq:mean_height}
\end{equation}
The resulting height field forms the basis for slope and roughness estimation. The grid resolution $d_{\mathrm{res}}$ reflects
terrain variations at the scale of the landing and support footprints.

\noindent
\textbf{Slope Estimation.}
The local slope at cell $g$ is computed from the normalized surface normal $\mathbf{n}(g)$ estimated via PCA over a neighborhood $\mathcal{N}_k(g)$. Let $\bar{\mathbf{p}}$ denote the centroid of neighboring points $\{\mathbf{p}_{g'}\}_{g'\in\mathcal{N}_k(g)}$. The covariance matrix is
\begin{equation}
\mathbf{C}_{\mathrm{cov}}(g)
=
\frac{1}{|\mathcal{N}_k(g)|}
\sum_{g' \in \mathcal{N}_k(g)}
(\mathbf{p}_{g'}-\bar{\mathbf{p}})
(\mathbf{p}_{g'}-\bar{\mathbf{p}})^{\top}.
\label{eq:pca_cov}
\end{equation}
Let $\mathbf{v}_0$ be the eigenvector associated with the smallest eigenvalue. The surface normal is $\mathbf{n}(g)=\mathbf{v}_0$ (oriented such that $n_z(g)\ge0$), and the slope is defined as
$s(g)=\arccos(n_z(g))$.
Slope captures local terrain inclination and is evaluated over a neighborhood size consistent with the support footprint.

\noindent
\textbf{Roughness Estimation.}
To capture larger-scale surface irregularities, we compute the height variation over a broader neighborhood $\mathcal{N}_{k'}(g)$ with $k'>k$. The neighborhood mean height is
\begin{equation}
\mu_h(g)
=
\frac{1}{|\mathcal{N}_{k'}(g)|}
\sum_{g' \in \mathcal{N}_{k'}(g)}
\bar{h}(g'),
\label{eq:mu_h}
\end{equation}
and roughness is the corresponding standard deviation
\begin{equation}
\sigma_h(g)
=
\sqrt{
\frac{1}{|\mathcal{N}_{k'}(g)|}
\sum_{g' \in \mathcal{N}_{k'}(g)}
(\bar{h}(g')-\mu_h(g))^2 },
\label{eq:sigma_h}
\end{equation}
capturing terrain unevenness beyond the local plane fit.

\noindent
\textbf{Clearance Estimation.}
Obstacle proximity is modeled through a clearance distance. Let $\mathcal{O}\subset\mathcal{G}$ denote occupied cells derived from semantic or geometric constraints. The clearance of cell $g$ is
\begin{equation}
d_{\mathrm{clear}}(g)
=
\min_{g' \in \mathcal{O}}
\|\mathbf{p}^{xy}_g - \mathbf{p}^{xy}_{g'}\|_2,
\label{eq:clearance}
\end{equation}
where $\mathbf{p}^{xy}_g=(x_g,y_g)$. This ensures sufficient separation from nearby obstacles for safe deployment.

\subsubsection{Geometric–Semantic Fusion Traversability Estimation}

Using the geometric descriptors and semantic attributes defined above,
we compute a continuous traversability score $T(g) \in [0,1]$ for each grid cell representing terrain suitability for UAV landing and UGV deployment. We first compute a geometric traversability score by combining slope, roughness, and clearance terms
\begin{equation}
T_{\mathrm{geo}}(g)
=
w_s T_{\mathrm{slope}}(g)
+
w_r T_{\mathrm{rough}}(g)
+
w_c T_{\mathrm{clear}}(g),
\end{equation}
where $w_s + w_r + w_c = 1$. The individual components are defined using normalized linear penalties
\begin{equation}
\begin{aligned}
T_{\mathrm{slope}}(g) &=
\mathrm{clip}\!\left(\frac{s_{\mathrm{hard}}-s(g)}
{s_{\mathrm{hard}}-s_{\mathrm{soft}}},0,1\right),\\
T_{\mathrm{rough}}(g) &=
\mathrm{clip}\!\left(\frac{\sigma_{\mathrm{hard}}-\sigma_h(g)}
{\sigma_{\mathrm{hard}}-\sigma_{\mathrm{soft}}},0,1\right),\\
T_{\mathrm{clear}}(g) &=
\mathrm{clip}\!\left(\frac{d_{\mathrm{clear}}(g)-d_{\mathrm{hard}}}
{d_{\mathrm{soft}}-d_{\mathrm{hard}}},0,1\right),
\end{aligned}
\end{equation}
where $s(g)$ is the terrain slope, $\sigma_h(g)$ denotes terrain roughness, and $d_{\mathrm{clear}}(g)$ is the obstacle clearance. The parameters $s_{\mathrm{soft}}, s_{\mathrm{hard}}, \sigma_{\mathrm{soft}}, \sigma_{\mathrm{hard}}, d_{\mathrm{hard}}, d_{\mathrm{soft}}$ define soft and hard thresholds controlling the transition between safe and unsafe terrain, and $\mathrm{clip}(x,0,1)=\min(\max(x,0),1)$. The thresholds reflect physical deployment limits on landing inclination, support-footprint height variation, and obstacle clearance. Higher normalized component scores indicate safer terrain. To incorporate terrain category information, we assign a semantic compatibility score.

\begin{equation}
T_{\mathrm{sem}}(g)=\tau(\hat{c}_{\mathrm{sem}}(g)),
\end{equation}
where $\tau(\cdot)$ maps each semantic class to a compatibility score in $[0,1]$, assigning high values to traversable terrain and zero to non-traversable classes (e.g., rock, gravel, water, etc.). Finally, geometric and semantic information are fused to produce the final traversability score
\begin{equation}
T(g)
=
\left(
\alpha T_{\mathrm{geo}}(g)
+
(1-\alpha) T_{\mathrm{sem}}(g)
\right)
p_{\mathrm{conf}}(g),
\end{equation}
where $\alpha \in [0,1]$ balances geometric and semantic contributions. The resulting traversability map provides a continuous spatial representation used for deployment-zone evaluation.
\subsubsection{Deployment Zone Selection}
Given the traversability map $T(g)$, candidate deployment locations are selected from grid cells satisfying $T(g) \ge T_{\mathrm{th}}$ and located within the maximum search radius from the target region.
Candidate cells are ranked by a distance-aware traversability objective
\begin{equation}
J(g)
=
(1-\lambda)T(g)
+
\lambda\left(
1-\frac{d(g,g_{\mathrm{goal}})}{r_{\max}}
\right),
\end{equation}
where $\lambda\in[0,1]$ balances traversability and proximity,
$d(g,g_{\mathrm{goal}})$ denotes Euclidean distance to the target
region, and $r_{\max}$ is maximum search radius. $J(g)$ favors safe landing locations closer to the target, encouraging short path traversal for the miniUGV. The top $K$ locations satisfying a
minimum separation constraint are selected as deployment targets.

\begin{figure}[t]
  \centering
  \includegraphics[width=0.90\linewidth]{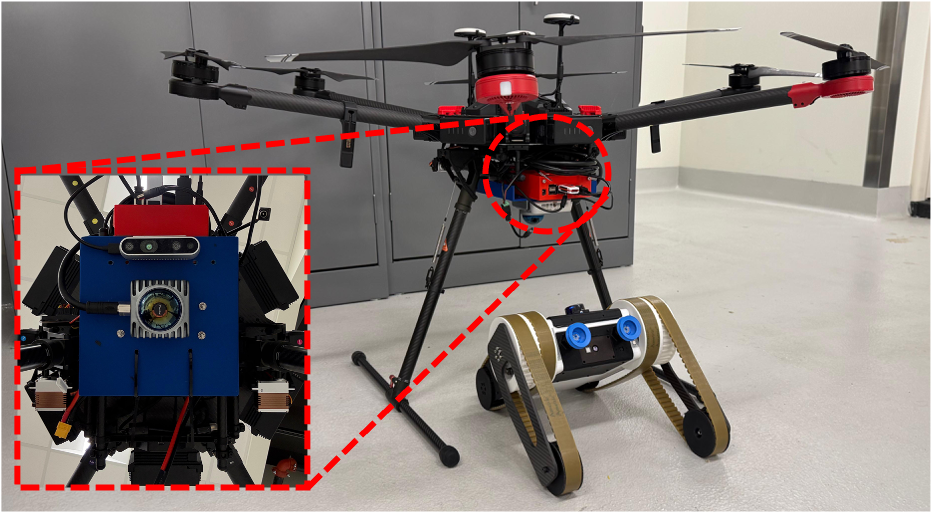}
  \caption{\small \textbf{Integrated aerial–ground robotic system}. The DJI Matrice M600 hexacopter carries a compact tracked ground robot and integrates a Livox Mid-360 LiDAR, RealSense camera, and Jetson Xavier NX compute module. FAST-LIO provides egomotion, but LiDAR point clouds are not used for reconstruction or traversability. \vspace{-10pt}}
  \label{fig:hardware_platform}
\end{figure}

\section{Experiments and Results}

\subsection{Dataset Collection}
We evaluate our method on high-resolution inspection videos collected during UAV field deployments targeting hidden spaces, using the platform shown in Fig.~\ref{fig:hardware_platform}. The platform is designed to perform multiple tasks during hidden-space inspection and therefore includes additional sensors such as LiDAR. However, our deployment analysis relies solely on RGB observations and onboard egomotion from FAST-LIO, and does not require LiDAR dense mapping. The dataset includes hidden-space scenarios such as culvert entrances embedded in natural terrain and ventilation systems partially occluded by surrounding structures. Each episode consists of a UAV surveying potential deployment regions using onboard RGB cameras prior to ground robot deployment. All observations are captured exclusively from top-down aerial views without oblique or ground-level imagery, reflecting realistic inspection conditions, unlike conventional reconstruction datasets with diverse viewing angles. Video sequences of 30 to 80 frames are acquired at 3 to 8 flight elevations ranging from low (10 to 20m) to high-altitude observations around 70m, per scene. Increasing altitude introduces reduced image resolution, limited parallax, and diminished geometric cues, making reconstruction progressively more challenging.

\definecolor{best}{RGB}{255,200,200}
\definecolor{second}{RGB}{255,230,170}
\definecolor{third}{RGB}{255,255,190}

\begin{table}
  \centering
  \small
  \begin{tabular}{@{}lcccc@{}}
    \toprule
    \textbf{\makecell{Model}} & \textbf{\makecell{ATE $\downarrow$}} & \textbf{\makecell{RPE\textsubscript{trans} $\downarrow$}} & \textbf{\makecell{RPE\textsubscript{rot} $\downarrow$}} \\
    \midrule
    LSM / CIPHER &  0.225 & 0.362 & 2.320  \\
    Easi3R &  0.481 & 0.422 & 2.901 \\
    Fast3R &  0.361 & 0.377 & \cellcolor{third}1.938  \\
    MASt3R &  \cellcolor{second}0.207 & \cellcolor{third}0.358 & 2.030  \\
    Ours & \cellcolor{third}0.222 & \cellcolor{second}0.338 & \cellcolor{second}1.920  \\
    \textbf{Ours + Relative Motion} & \cellcolor{best}\textbf{0.194} & \cellcolor{best}\textbf{0.310} & \cellcolor{best}\textbf{1.878} \\
    \bottomrule
  \end{tabular}
  \caption{\small \textbf{Comparison of Multi-view Pose Estimation.}
  }
  \label{tab:performance_metrics}
\end{table}

\begin{table}
  \centering
  \small
  \begin{tabular}{@{}lcc@{\hspace{6pt}}}
    \toprule
    \textbf{\makecell{Model}} & \textbf{\makecell{Peak GPU \\ Memory (GB) $\downarrow$}} & \textbf{\makecell{Inference \\ Time (sec) $\downarrow$}} \\
    \midrule
    LSM / CIPHER  &  28.89 &  526.47 \\
    Easi3R & 33.26 & 233.61  \\
    Fast3R & \cellcolor{third}19.24 & \cellcolor{third}7.60  \\
    MASt3R & 24.42 &  15.20   \\
    \textbf{Ours} & \textbf{\cellcolor{best}14.09} & \cellcolor{best}\textbf{3.37}  \\
    Ours + Relative Motion &  \cellcolor{second}16.41 & \cellcolor{second}5.28   \\
    \bottomrule
  \end{tabular}
  \caption{\small \textbf{Comparison of Average Inference Efficiency.} Runtime was measured with one RTX A6000 GPU.
  }
  \label{tab:efficiency_metrics}\vspace{-10pt}
\end{table}

\begin{figure*}[t]
  \centering
  \includegraphics[width=0.95\linewidth]{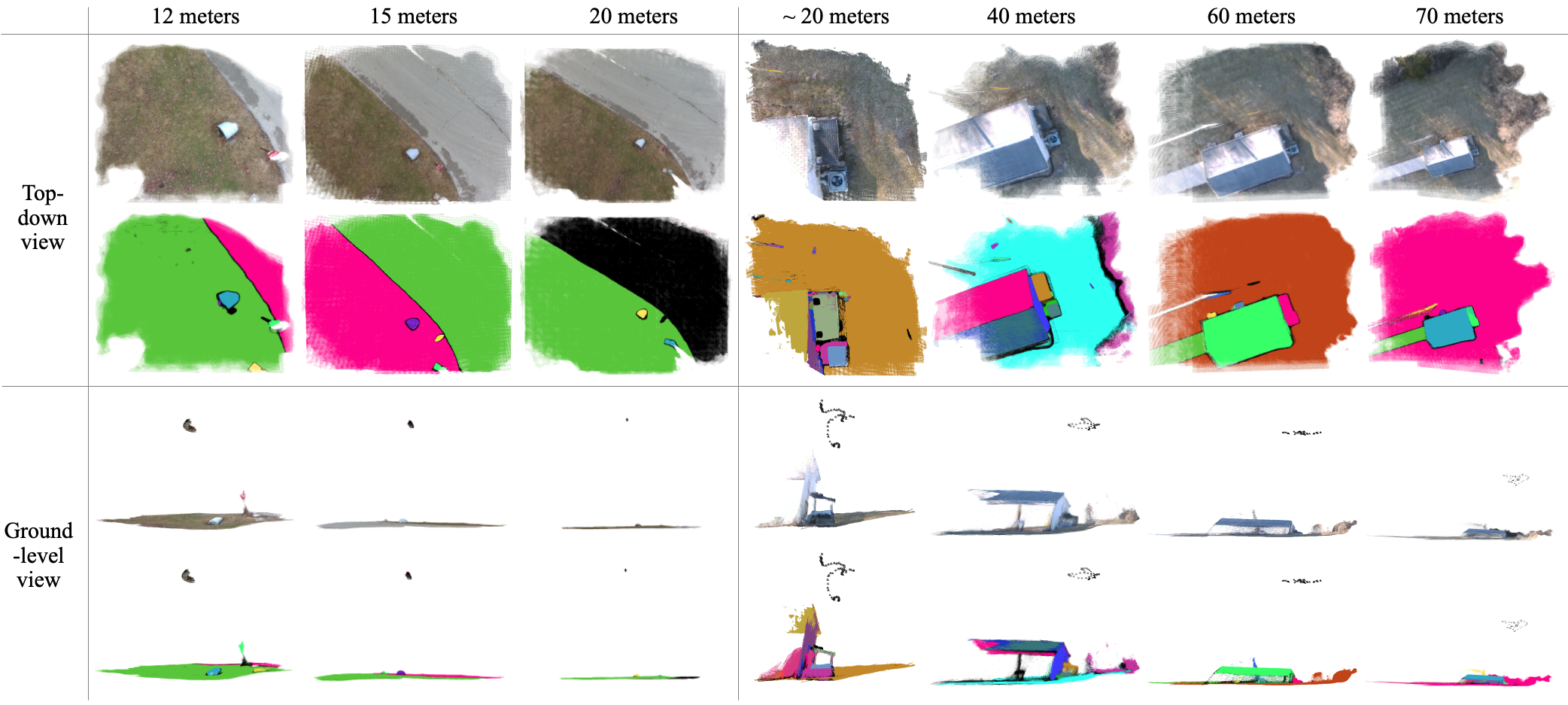}
  \caption{\small \textbf{RGB-based geometric and semantic 3D reconstruction from purely top-down aerial views at increasing UAV elevations}: a culvert entrance (left) and a vent system occluded by surrounding structures (right). For each elevation, RGB reconstructions are shown above their corresponding semantic reconstructions for both top-down and ground-level views. Ground-level views also include the camera trajectory. Despite limited viewpoint parallax for overhead observations and higher altitudes, our method maintains geometrically consistent ground-level renderings and accurately segments deployment-relevant targets  across a wide range of flight altitudes from RGB imagery.
  Semantic mask granularity can be adjusted to capture structures at different spatial scales as required. \vspace{-10pt}}
  \label{fig:elevations}
\end{figure*}

\begin{figure*}[t]
  \centering
  \includegraphics[width=0.85\linewidth, trim=13 0cm 0 0cm, clip]{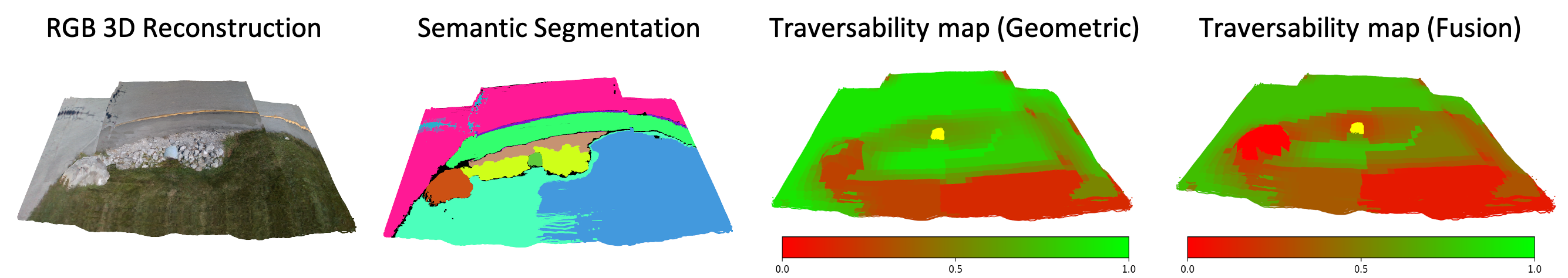}
  \caption{\small \textbf{Visual Results of Traversability mapping.} From left to right: RGB-based 3D reconstruction, semantic segmentation, geometric traversability map, and geometric–semantic fusion result. The geometric map assigns high scores to planar regions but partially misclassifies rocky areas due to smoothed geometry. The fusion map assigns low traversability to terrain classes such as rocks and structural obstacles, producing a more conservative and deployment-consistent representation. \vspace{-10pt}}
  \label{fig:traver_results}
\end{figure*}

\subsection{Evaluation of 3D Reconstruction}

Following the evaluation protocol of~\cite{cong2025e3d}, we compare representative multi-view reconstruction baselines in pose estimation accuracy, input scalability, inference runtime, and peak GPU memory usage. All methods are evaluated zero-shot on unseen field scenes without scene-specific fine-tuning to demonstrate generalizability, and our method is further assessed qualitatively for cross-view consistency across flight elevations.
We select recent efficient or lightweight multi-view models including LSM/CIPHER~\cite{fan2024large,lee2025cipher}, Easi3R~\cite{chen2025easi3r}, Fast3R~\cite{yang2025fast3r}, and MASt3R~\cite{leroy2024grounding}. While 3D Gaussian Splatting based works often achieve high rendering fidelity, we exclude them since they require scene-specific test-time optimization and do not generalize zero-shot across scenes. LSM, CIPHER and MASt3R are efficient but operate on only two input images, making reconstruction sensitive to input selection and limiting scalability to longer input sequences. To enable multi-view comparison, we extend these pairwise models by building a scene graph over all input, running the pairwise model on each image pair, and recovering a single trajectory via global point-cloud alignment.

\noindent
\textbf{Pose Estimation \& Efficiency.}
Tables~\ref{tab:performance_metrics} and \ref{tab:efficiency_metrics} summarize pose estimation accuracy and efficiency metrics measured on a single RTX A6000 GPU. Since most RGB-based reconstruction models estimate poses up to an unknown global scale, predicted trajectories are aligned to ground truth using closed-form Sim(3) Umeyama alignment. We report Absolute Trajectory Error (ATE), Relative Translation Error (RPE\textsubscript{trans}), and Relative Rotation Error (RPE\textsubscript{rot}), where ATE measures global trajectory consistency and RPE metrics evaluate frame-to-frame motion accuracy.
Our methods achieve competitive peak GPU memory and fastest inference times while maintaining superior pose accuracy and scaling to arbitrary-length sequences
without per-scene optimization.

\noindent
\textbf{Reconstruction at Different Altitudes.}
We evaluate the robustness of RGB-based geometric and semantic reconstruction under purely top-down aerial observations collected at varying UAV elevations. This setting is challenging for vision-based geometry estimation, as limited triangulation baselines and weak perspective variation lead to depth ambiguity and unstable pose estimation. The difficulty further increases with altitude as scene structures occupy fewer pixels and geometric cues diminish.

Figure~\ref{fig:elevations} presents qualitative results from two hidden-space inspection scenarios: a culvert entry and a vent system. For each scene, geometric and corresponding semantic reconstructions are generated using RGB footage captured solely from top-down views across elevations ranging from low to high altitudes exceeding 20 and 70 meters, respectively. Despite the absence of oblique or ground-level observations, the proposed RGB-based framework consistently recovers meaningful ground-level geometry and structural layout across all elevations. Simultaneously, accurate semantic segmentation of deployment targets, such as culvert openings and ventilation structures, is maintained, enabling reliable identification of candidate deployment regions even from distant aerial viewpoints. Segmentation granularity can also be adjusted to meet varying deployment precision requirements.

This robustness is critical for aerial-to-ground deployment, where UAV operation is often restricted to overhead views for safety or accessibility constraints. Unlike depth sensors such as LiDAR, whose accuracy degrades significantly with sensing distance and grazing angles, the learned RGB-based reconstructions remain stable even at high altitudes. These results indicate metrically grounded RGB-based reconstruction enables reliable terrain understanding for traversability reasoning
without requiring heavy sensing payloads.

\subsection{Evaluation of Traversability}

We evaluate the proposed traversability framework through both qualitative visualization and quantitative analysis using real-world aerial datasets collected from our UAV platform. Figure~\ref{fig:traver_results} presents representative results including the RGB-based 3D reconstruction, semantic segmentation, geometric traversability map, and the geometric–semantic fusion map. The geometric traversability map assigns high scores to large planar regions such as pavement and flat grass while suppressing steep embankments and visibly uneven terrain. However, limitations arise near terrain boundaries containing small rocks and debris. Due to some geometric imprecision of the reconstructed point cloud, fine-scale surface irregularities are partially smoothed, causing some physically unstable regions to appear moderately traversable. Consequently, certain locations unsuitable for safe UAV landing or UGV deployment may be incorrectly identified as viable deployment zones.
\begin{table}
  \centering
  \small
  \begin{tabular}{@{}lcccc@{}}
    \toprule
    \textbf{\makecell{Model}} & \textbf{\makecell{mACC $\uparrow$}} & \textbf{\makecell{aAcc $\uparrow$}} & \textbf{\makecell{ROC AUC $\uparrow$}} & \textbf{\makecell{MSE $\downarrow$}}\\
    \midrule
    Geometric & 0.779 & 0.841 & 0.926 & 0.113 \\
    \textbf{Fusion} & \textbf{0.958} & \textbf{0.938} & \textbf{0.948} & \textbf{0.042}  \\
    \bottomrule
  \end{tabular}
  \caption{\small \textbf{Comparison of Traversability Accuracy.}
  }
  \label{tab:traversability_metrics} \vspace{-20pt}
\end{table}
The geometric–semantic fusion map alleviates these ambiguities by incorporating semantic information. Terrain categories associated with hazardous conditions, such as rocks and structural obstacles, are consistently assigned low traversability even when local geometric cues appear acceptable. As a result, the fused representation produces more spatially consistent terrain classification and effectively suppresses semantically hazardous regions that could otherwise lead to unsafe deployment decisions.

Quantitative results are summarized in Table~\ref{tab:traversability_metrics}. Ground-truth annotations are obtained by labeling BEV grid cells according to terrain semantics and deployment safety criteria, where flat obstacle-free support regions are treated as traversable and cells containing rocks, steep terrain, structural obstacles, or insufficient clearance are treated as non-traversable. Traversability scores are thresholded at 0.5 to produce binary labels for comparison. The geometric–semantic fusion approach significantly improves performance across all metrics, with mean accuracy, overall accuracy, and ROC AUC enhanced while mean square error decreased, indicating more reliable traversability estimation. These results demonstrate that integrating semantic information effectively compensates for limitations of geometric reconstruction and yields more reliable traversability predictions for deployment-zone selection.

\begin{figure}[t]
  \centering
  \includegraphics[width=0.9\linewidth, trim=0 1.3cm 0 0.3cm, clip
  ]{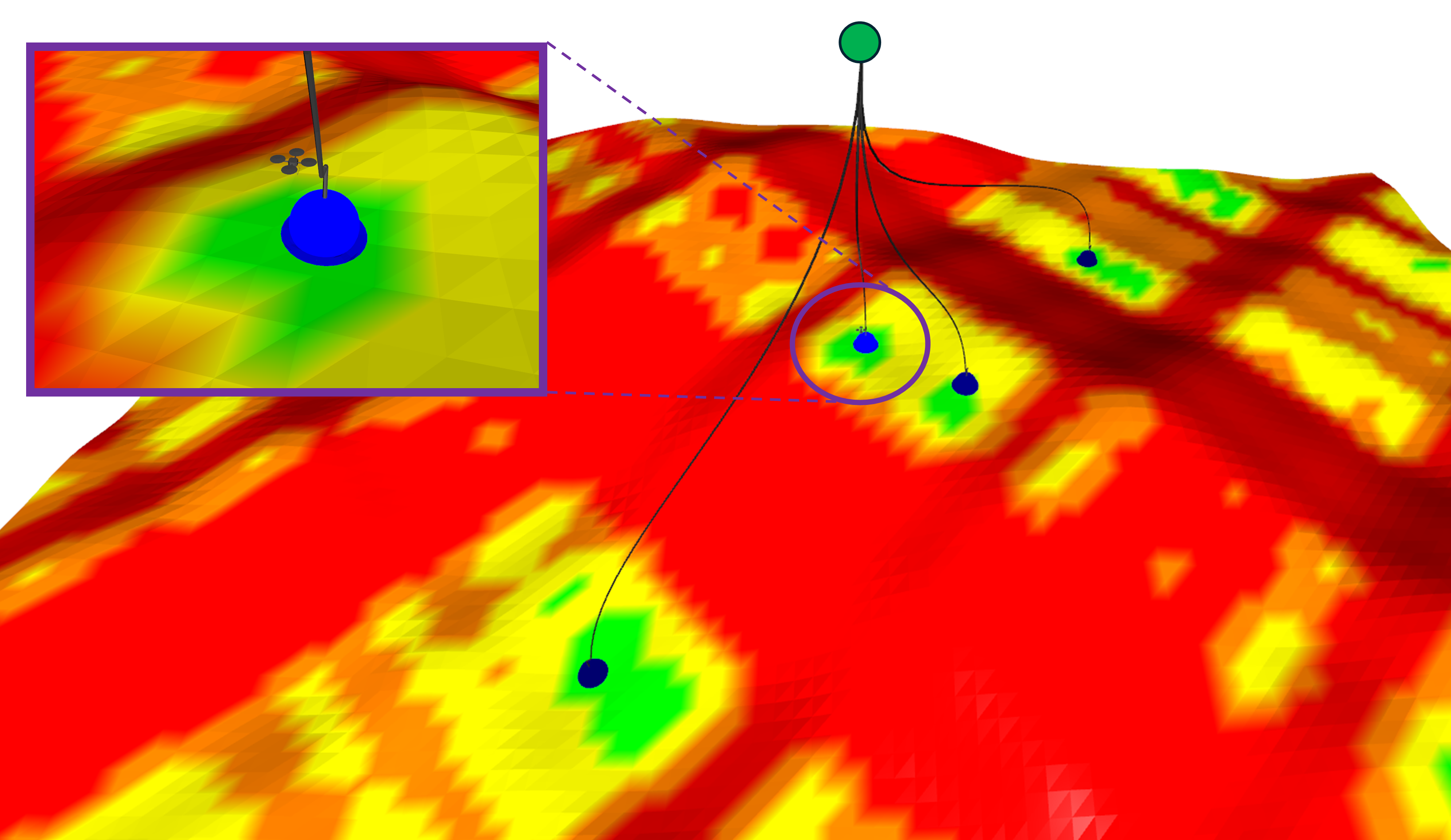}
  \caption{\small \textbf{Deployment location selection and trajectory generation.} High traversability regions (green) indicate safe deployment areas, while red regions denote unsafe terrain. Candidate deployment locations (blue) are identified and feasible trajectories are generated from the UAV to the deployment zone.\vspace{-10pt}}
  \label{fig:path_plan}
\end{figure}

\subsection{Deployment and Trajectory Generation}

To validate the deployability of the proposed aerial traversability representation, we generate dynamically feasible trajectories from the UAV's current state to candidate deployment zones using the hazard-aware landing optimization (HALO) framework~\cite{hayner2023halohazardawarelandingoptimization}. The planner operates on the deployment regions extracted from the traversability map and produces feasible trajectories toward selected deployment targets under simplified UAV dynamics constraints. Figure ~\ref{fig:path_plan} illustrates regions with high traversability (green),
candidate deployment locations (blue), and the resulting trajectory generated by the planner. The trajectory remains within high-traversability regions and avoids terrain identified as unsafe in the fused map, indicating that the selected deployment zones are both locally safe and dynamically reachable. The planned trajectory is executed using a low-level PID controller that tracks the reference path and guides the UAV to the final deployment location.

\section{Conclusion and Future Work}

We presented a metric RGB-based traversability analysis framework for aerial-to-ground deployment in hidden-space inspection. Under purely top-down UAV observations, the system combines feed-forward multi-view RGB-based reconstruction, temporally consistent semantic mapping, and metric-scale grounding using onboard egomotion. The resulting geometric–semantic map enables confidence-aware BEV traversability estimation that fuses geometric features, semantic compatibility, and confidence. Experiments on real-world aerial inspection data demonstrate competitive pose accuracy, computational efficiency, zero-shot generalization, cross-view consistency, and deployment-consistent terrain assessment. Semantic–geometric fusion suppresses ambiguous yet hazardous regions and supports curvature-constrained path validation, indicating that selected deployment zones are both locally safe and globally reachable.
Full end-to-end deployment, including physical UGV release, post-deployment traversal, recovery, and safety-aware failure handling, remains an important direction for future work.

\section*{Acknowledgments}
Z. Wang is supported by DARPA ANSR (RTX CW2231110), DARPA TIAMAT (HR0011-24-9-0431), ARL StAmant (W911NF-23-S-0001), as well as the NSF AI Institute for Foundations of Machine Learning (IFML).

\bibliographystyle{IEEEtran}
\bibliography{refs}

\end{document}